\documentclass[11pt]{article}

\usepackage[final]{acl}

\usepackage{times}
\usepackage{latexsym}
\usepackage{amsmath}
\usepackage{mathtools} 

\usepackage[T1]{fontenc}

\usepackage[utf8]{inputenc}

\usepackage{microtype}

\usepackage{inconsolata}

\usepackage{graphicx}
\usepackage{amssymb}
\usepackage{multirow}
\usepackage{booktabs}
\usepackage{tcolorbox}
\tcbuselibrary{skins, breakable}
\usepackage{enumitem} 
\usepackage{caption}

\usepackage{listings}        
\usepackage{xcolor}          

\lstdefinestyle{jsonstyle}{
    basicstyle=\ttfamily\small,
    columns=fullflexible,
    breaklines=true,
    captionpos=b,
    frame=single,
    framerule=0pt, 
    backgroundcolor=\color{white}, 
    keywords={Action, Objects},
    keywordstyle=\bfseries\color{blue!60!black},
    stringstyle=\color{green!40!black},
    commentstyle=\color{gray},
    showstringspaces=false,
    keepspaces=true,
}

\title{Enhancing Agentic Textual Graph Retrieval with Synthetic Stepwise Supervision}

\author{
  \textbf{Ge Chang}\textsuperscript{1,2,$\ast$,$\ddagger$}, 
  \textbf{Jinbo Su}\textsuperscript{1,3,$\ast$,$\ddagger$}, 
  \textbf{Jiacheng Liu}\textsuperscript{1,2,$\ast$,$\ddagger$}, 
  \textbf{Pengfei Yang}\textsuperscript{1,4,$\ddagger$}, 
  \textbf{Yuhao Shang}\textsuperscript{1,5,$\ddagger$}, 
\\
  \textbf{Huiwen Zheng}\textsuperscript{6}, 
  \textbf{Hongli Ma}\textsuperscript{6}, 
  \textbf{Yan Liang}\textsuperscript{6}, 
  \textbf{Yuanchun Li}\textsuperscript{1,$\dagger$}, 
  \textbf{Yunxin Liu}\textsuperscript{1} 
\\
\\
  \textsuperscript{1}Institute for AI Industry Research (AIR), Tsinghua University \\
  \textsuperscript{2}Peking University \quad
  \textsuperscript{3}School of Information, Renmin University of China \\
  \textsuperscript{4}Harbin Institute of Technology \quad
  \textsuperscript{5}North China Electric Power University \\
  \textsuperscript{6}GDS Holdings Limited 
\\
  \noalign{\vskip 0.1in}
  \texttt{\{gchang25, jiachengliu25\}@stu.pku.edu.cn}, \texttt{sujinbo@ruc.edu.cn} \\
  \texttt{liyuanchun@air.tsinghua.edu.cn}
}

\begin{document}
\maketitle
{
  \renewcommand{\thefootnote}{\fnsymbol{footnote}}
  \footnotetext[1]{Equal contribution.}
  \footnotetext[2]{Corresponding author.}
  \footnotetext[3]{Work done during internship at AIR, Tsinghua University.}
  \renewcommand{\thefootnote}{\arabic{footnote}}
}
\begin{abstract}
Integrating textual graphs into Large Language Models (LLMs) is promising for complex graph-based QA. However, a key bottleneck is retrieving informative yet compact subgraphs that fit the LLM context. Existing retrievers often struggle, relying either on shallow embedding similarity or costly interactive policies that require excessive supervision.
To address these challenges, we introduce Graph-S$^3$, an agentic textual graph reasoning framework featuring an LLM-based retriever trained with synthetic stepwise supervision. Rather than relying on final answer rewards—which often yield sparse and unstable signals—we optimize the retriever by evaluating each step against offline-extracted golden subgraphs.
Our approach distills golden subgraphs via a specialized data synthesis pipeline to formulate dense rewards, facilitating a two-stage training scheme that effectively learns the interactive graph exploration policy.
Based on extensive experiments on three common datasets in comparison with several strong baselines, our approach achieves an average improvement of 15.6\% in accuracy and 17.2\% in F$_1$ score. The advantage is even higher in more complicated multi-hop reasoning tasks. 

\end{abstract}

\section{Introduction}

Textual graphs are widely employed for structured knowledge representation in domains like question answering and scientific discovery~\citep{peng2024graph, procko2024graph, zhang2025survey}. By explicitly modeling multi-hop relations, they enable interpretable, compositional reasoning that is difficult to achieve with unstructured text~\citep{chen2020review, hogan2021knowledge,zou2020survey}. While early methods were constrained by inflexible symbolic inference and high annotation costs~\citep{yih2016value}, the semantic capabilities of Large Language Models (LLMs) have alleviated these limitations~\citep{chang2024survey,su2025racqc}. Consequently, integrating LLMs with textual graphs has become a prominent paradigm for general-purpose graph understanding and QA~\citep{lewis2020retrieval, peng2024graph, procko2024graph, zhang2025survey, jin2024large, chai2023graphllm}.

\begin{figure*}[t]
  \includegraphics[width=0.98\linewidth]{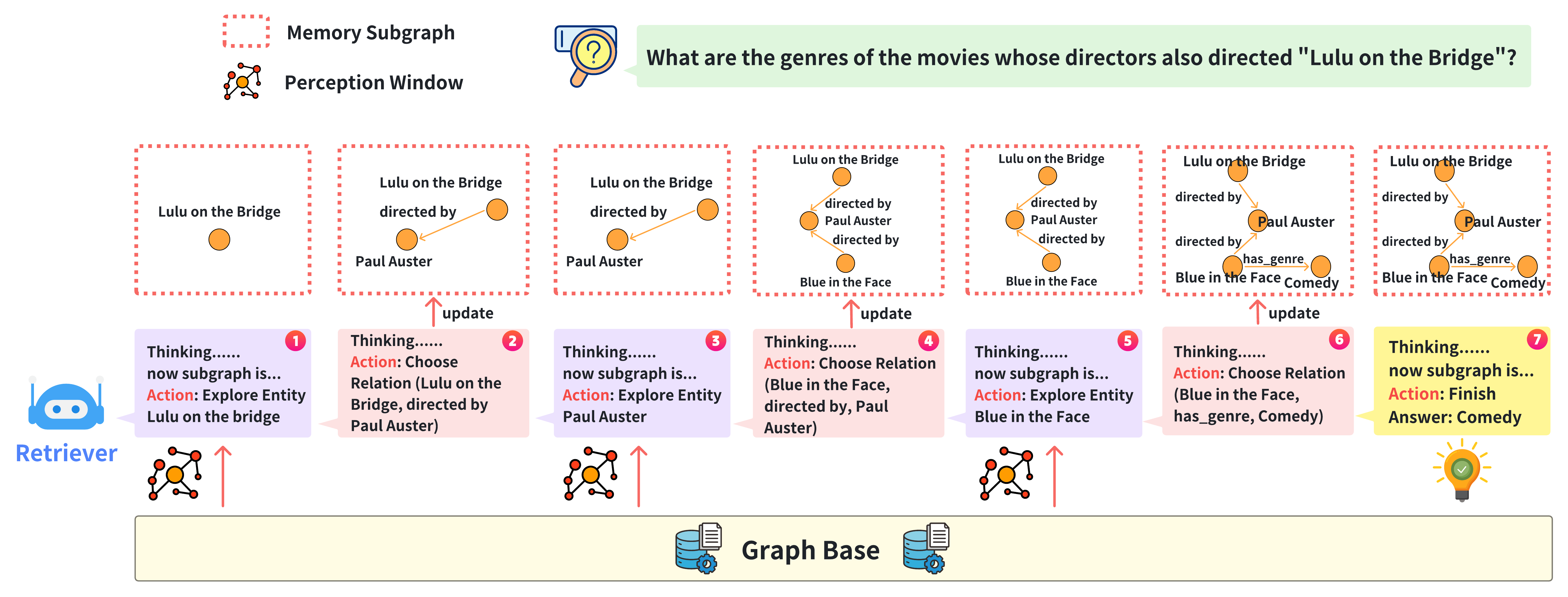} \hfill
  \caption {An illustration of agentic textual graph retrieval and question answering.}
  \label{fig1}
\end{figure*}



LLM agents accessing graphs via tool calls surpass simple retrieval~\citep{jiang2024kg,yang2024graphagent, ji2024retrieval,xu2026unlocking}, as illustrated in Figure~\ref{fig1} where an agent iteratively collects information. However, training such agents via supervised fine-tuning (SFT) often collapses the action space, favoring memorization over generalizable policies~\citep{chu2025sft,li2024entropic}. While outcome-supervised reinforcement learning (RL) offers a potential solution by exploring reasoning trajectories~\citep{lightman2023let,paolo2024discovering}, it faces significant hurdles in this domain: the action space in real-world textual graphs is often too vast for efficient exploration, and outcome-based rewards are sparse and noisy. For instance, redundant or erroneous retrieval steps may coincidentally lead to the correct final answer, rendering outcome supervision unreliable for assessing reasoning quality~\citep{liu2023lazy, rengarajan2022reinforcement}.


To overcome this limitation, we introduce a synthetic stepwise supervision scheme that provides explicit feedback at every decision step, ensuring that the model is guided not only by the correctness of the final answer but also by the quality of intermediate actions.
The key idea is to \textbf{guide each graph retrieval step with golden subgraphs offline extracted from the target graph}.
Specifically, we propose an automated pipeline to construct the golden subgraphs for reward computation. 
We first generate a large amount of subgraph candidates through random and LLM-guided exploration, and filter the candidates based on \textbf{information sufficiency}, i.e. whether they are able to produce correct answers with LLMs. These successful exploratory trajectories are used for SFT, providing the retriever with basic navigation ability as a warm-up stage. Then we further refine the subgraphs to enhance \textbf{information conciseness} by iteratively pruning redundant content while preserving answer consistency. With these refined subgraphs, each online graph retrieval action can be associated with an explicit stepwise reward based on its contribution to the golden subgraphs. The combined two-step training pipeline guides the retriever to improve reasoning decisions over long action chains.

Extensive experiments have demonstrated the effectiveness of our synthetic stepwise supervision approach. For example, while retrieving only 11.44\% of the triples, Graph-S$^3$ achieves an average improvement of 15.6\% in accuracy and 17.2\% in F$_1$ score across the WebQSP, CWQ, and MetaQA benchmarks.

\par
In summary, the main contributions of this work are as follows:
\par
(1) We propose an automatic pipeline for synthesizing high-quality stepwise supervision data for interactive graph retrieval, addressing the scarcity of fine-grained training signals in this field.
\par
(2) We design a two-stage training paradigm tailored for graph reasoning: SFT on raw synthetic trajectories to bootstrap basic navigation ability, followed by RL with synthetic stepwise rewards on refined trajectories to provide explicit feedback and strengthen reasoning strategies.
\par
(3) Experimental results demonstrate that Graph-S$^3$ achieves state-of-the-art performance on the WebQSP, CWQ, and MetaQA datasets with accurate and compact graph retrieval.

\begin{figure*}[t]
    \includegraphics[width=0.98\textwidth]{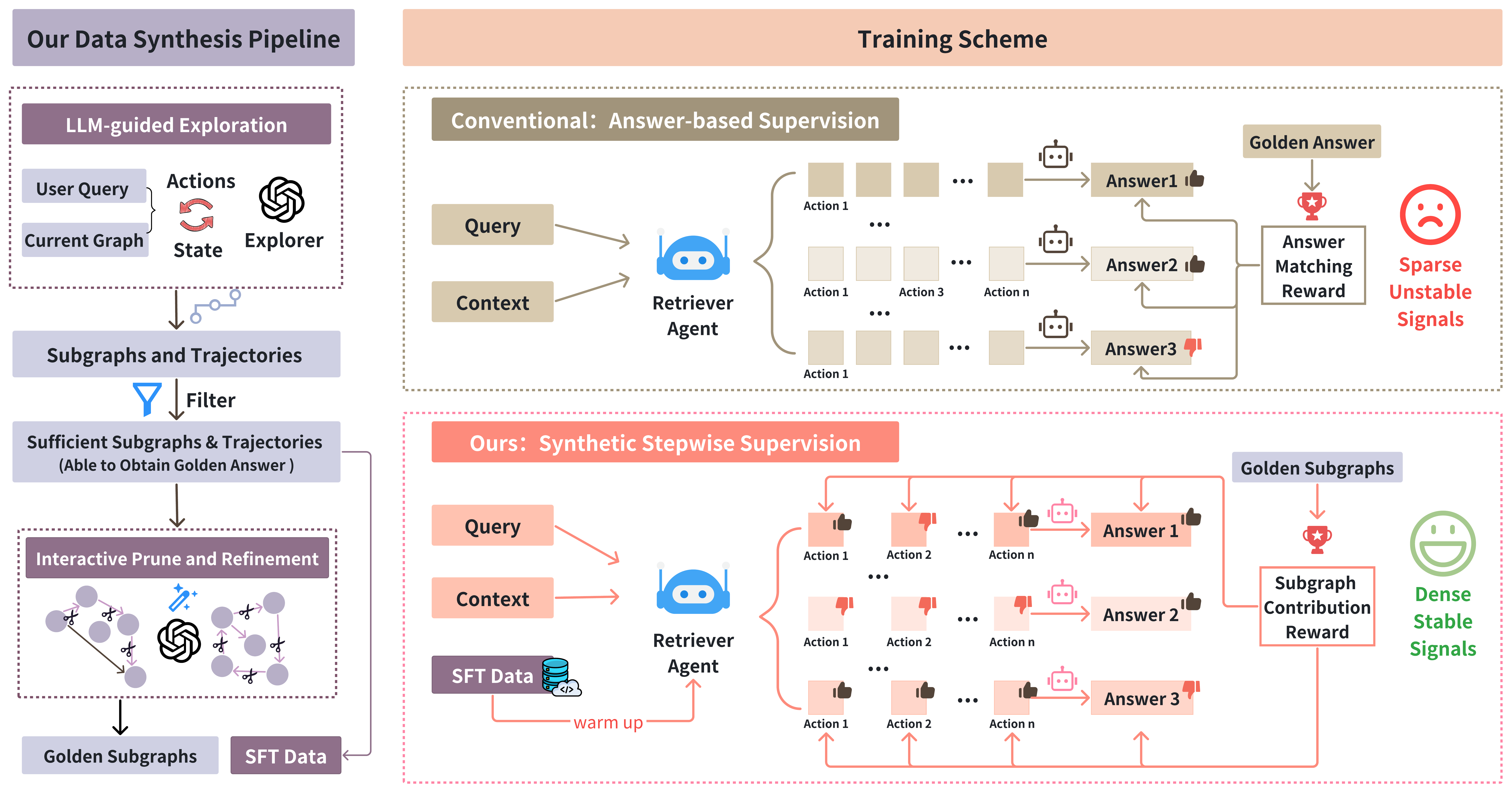} \hfill
    \caption{An Overview of our data synthesis pipeline and training scheme.}
    \label{fig:main_fig}
\end{figure*}

\section{Related Work}

\subsection{Graph Retrieval Methods}
Graph retrieval approaches include similarity-based, GNN-based, and LLM-based methods~\citep{peng2024graph, procko2024graph, zhang2025survey,zhu2025graph,han2024retrieval,zhou2026arkanswercentricretrievertuning}, but most perform one-shot retrieval and often return redundant or incomplete subgraphs.  
Recent interactive frameworks~\citep{jiang2024kg,ji2024retrieval,yang2024graphagent} allow iterative exploration, yet their training predominantly relies on imitation of language patterns or outcome-based supervision, which provides only coarse feedback and limits stable multi-hop reasoning.  
In contrast, our work employs RL with synthetic stepwise rewards and a scalable data synthesis pipeline to provide supervision for interactive retrieval.

\subsection{Stepwise RL for Graph Reasoning}
Recent advances such as OpenAI o1 and DeepSeek-R1~\citep{jaech2024openai, guo2025deepseek} demonstrate the effectiveness of RL in strengthening multi-step reasoning, enabling models to perform longer chains of thought with improved reliability in domains like mathematics and programming~\citep{guo2024deepseek,el2025competitive,li2026sesearchselfevolvingsearchagent,xu2025videosegr1reasoningvideoobjectsegmentation}. 
In contrast, applying RL to textual graphs remains limited, partly due to the lack of fine-grained supervision data~\citep{zhang2025survey, yao2025learning, LIU2025111625}. 
RL-based graph agents ~\citep{das2017go,cui2025process} relied on sparse outcome rewards, making credit assignment across reasoning trajectories difficult. Although recent efforts have introduced reasoning-structured datasets~\citep{pahilajani2024grs}, high-quality stepwise supervision for graph-based RL remains scarce and difficult to construct at scale. These observations highlight the need for scalable approaches that can provide fine-grained supervision signals and support stable optimization for interactive graph retrieval.

\section{Method}
\label{method}

We present our agentic retrieval framework, designed to equip large language models with robust graph reasoning capabilities through stepwise supervision, a two-stage training paradigm, and an interactive retrieval strategy. As illustrated in Figure~\ref{fig:main_fig}, the framework comprises three main components. First, we construct an automatic data synthesis pipeline that leverages GPT-4o~\citep{OpenAI_GPT4o_System_Card_2024} to generate diverse exploratory trajectories, which are subsequently refined into high-quality stepwise supervision data. This addresses the scarcity of fine-grained training signals for graph-based RL. Second, we adopt a two-stage training paradigm: SFT on raw synthetic trajectories provides a warm-up initialization for basic graph navigation, while RL with synthetic stepwise rewards on refined trajectories supplies explicit feedback at each decision step, stabilizing optimization and strengthening reasoning strategies. Finally, during inference, the retriever operates under an interactive retrieval mechanism that conducts stepwise, structure-aware exploration of the textual graph, thereby reducing redundancy and mitigating incomplete retrieval.

\subsection{Data Synthesis}
Existing LLMs are not pretrained on graph-structured data~\citep{zhang2025r1}, which significantly limits their performance on graph reasoning tasks. As a result, effective training requires high-quality supervision to cultivate graph comprehension and reasoning capabilities. However, constructing such datasets is notoriously expensive, since it often relies on manual annotation by domain experts~\citep{choubey2024distill}, leading to a persistent scarcity of high-quality graph reasoning data. To address this issue, we design a pipeline that automatically synthesizes graph reasoning trajectories. Specifically, we first define a set of actions that enable structured interaction with the graph. Then, we leverage GPT-4o to perform these actions and collect valid action–response pairs, which form exploratory trajectories. The raw trajectories are directly used for SFT to provide the model with basic navigation ability, while a refinement step prunes redundant detours and preserves all answer-consistent trajectories, producing high-quality stepwise supervision data for RL. 

\subsubsection{Action Space for Graph Exploration}
\label{subsec:ActionSpaceForGraphexploration}

Given a textual graph $\mathcal{G} = \{t_i\}_{i=1}^m$, where each triple 
$t_i = (e_h^i, r^i, e_t^i) \in \mathcal{E} \times \mathcal{R} \times \mathcal{E}$ 
consists of a head entity $e_h^i \in \mathcal{E}$, a relation $r^i \in \mathcal{R}$, 
and a tail entity $e_t^i \in \mathcal{E}$. Here, $\mathcal{E}$ and $\mathcal{R}$ 
denote the sets of entities and relations, respectively. To enable stepwise exploration over $\mathcal{G}$, we define the retriever’s action space as consisting of three types of operations. For clarity, we use $(x,r,y)$ to denote a generic triple in $\mathcal{G}$, 
where $x,y \in \mathcal{E}$ and $r \in \mathcal{R}$.

\textbf{Explore Entity}: This operation expands the local neighborhood of a given entity by retrieving all directly connected triples in $\mathcal{G}$. Formally, for a target entity $x \in \mathcal{E}$, the operation is defined as
\begin{align}\label{eq:explore_entity}
    \text{Explore}(x) = & \{(x,r,y) | (x,r,y) \in \mathcal{G}^{p}\} \nonumber \\
                       & \cup \{(y,r,x) | (y,r,x) \in \mathcal{G}^{p}\}
\end{align}
where $r \in \mathcal{R}$ and $y \in \mathcal{E}$ denote relations and neighboring entities, respectively. The retrieved triples are added to the perception window $\mathcal{G}^{p}$ for subsequent reasoning steps.

\textbf{Choose Relation}: The perception window $\mathcal{G}^{p}$ obtained from the 
\textsc{Explore} action may still contain many irrelevant triples. To avoid 
introducing redundant context into the LLM, this operation prunes $\mathcal{G}^{p}$ 
into a query-relevant subgraph $\mathcal{G}^{sub}$: 
\begin{equation}\label{eq:choose}
\begin{split}
    \operatorname{Choose}(q,\mathcal G^{p}) = \bigl\{ &(x,r,y) \in \mathcal G^{p} \big| \\
    & F(q,(x,r,y)) = 1 \bigr\}.
\end{split}
\end{equation}
Here, $q$ is the query, $(x,r,y)$ denotes a triple in $\mathcal{G}^{p}$, and 
$F(q,(x,r,y))$ is the relevance function learned by the retriever to decide 
whether the triple should be preserved. $F$ outputs $1$ if the triple is judged 
relevant to the query and $0$ otherwise. 

\textbf{Finish}: This operation indicates that the retriever has gathered sufficient 
evidence in the current subgraph $\mathcal{G}^{sub}$ to answer the query $q$. 
Once invoked, the exploration process terminates, and $(q,\mathcal{G}^{sub})$ is 
used to produce the final answer:
\begin{align}\label{eq:stop}
   \text{Finish}(q,\mathcal{G}^{sub}) = \text{Answer}(q,\mathcal{G}^{sub}),
\end{align}
where $\text{Answer}(q,\mathcal{G}^{sub})$ denotes answering query $q$ based on 
the retrieved subgraph.

\subsubsection{Graph Reasoning Data Synthesis}
\label{subsubsec:Graph_Reasoning_Data_Synthesis}

Given the defined action space, we synthesize reasoning trajectories by letting a behavior model (GPT-4o) interact with the graph. We formalize the generation process as a Markov decision process (MDP) with deterministic transitions defined by the graph action space. Each trajectory consists of multiple decision steps, and is later decomposed into step-level training instances for SFT and RL.

\textbf{State.}
We define each state as a tuple of four components: $s_t = \big(q,\; \mathcal G^p_t,\; \mathcal G^{sub}_t,\; h_t\big)$,
where $q$ is the query, $\mathcal G^p_t$ the perception window aggregated by \textsc{Explore}, and $\mathcal G^{sub}_t$ the focused subgraph maintained by \textsc{Choose}, and $h_t = (a_1,\dots,a_{t-1})$ the action history up to step $t$. Including $h_t$ allows the retriever to condition its decisions not only on the current graph view but also on the reasoning trajectory already taken.

\textbf{Action.}
The parameterized action space is $\mathcal A$, defined in \ref{subsec:ActionSpaceForGraphexploration}.
During the synthesis phase, the behavior model selects $a_t\in\mathcal A$ given $s_t$, and additionally produces a natural language reasoning trace that explains the choice of action.

\textbf{Transition.} The execution of action $a_t$ triggers a state transition $s_t \to s_{t+1}$ by updating the action history $h_{t+1} \leftarrow h_t \cup \{a_t\}$ and modifying the graph components based on the action type. Specifically, if $a_t = \textsc{Explore}(x_t)$, the agent expands the partial graph $\mathcal{G}^p_t$ by merging it with the newly explored neighbors. If $a_t = \textsc{Choose}$, the agent updates the subgraph $\mathcal{G}^{sub}_t$ by filtering relevant triples from $\mathcal{G}^p_t$ via the selection function. The episode terminates when $a_t = \textsc{Finish}$ or the time step $t$ reaches the maximum limit $T_{\max}$.

\textbf{Answer and retention.}
Upon termination at step $T$, we produce an answer $\hat y\;=\;\text{Answer}\!\big(q,\mathcal G^{sub}_T\big)$
and keep the trajectory $\tau \;=\; \big\{(s_t,a_t)\big\}_{t=1}^{T}$
only if $\hat y$ matches the set of ground-truth answers. 
After that, we use the raw action labels for SFT dataset: $\mathcal D_{\text{SFT}} \;=\; \big\{(s_t,a_t)\big\}_{\tau,\,t}$.
In the next subsection, we further refine trajectories to obtain stepwise supervision signals for RL.

\subsubsection{Trajectory Refinement for RL}
\label{subsec:refinement}

While SFT directly benefits from raw trajectories, RL requires more concise training signals~\citep{yue2025does}. 
Filtering trajectories only by final answer correctness often produces redundant exploration steps, introducing noise and inefficiency during policy optimization. 
To address this, we introduce a refinement procedure that removes unnecessary detours while preserving all answer-consistent trajectories, thereby yielding shorter and cleaner trajectories for RL.

Let the set of retained raw trajectories be $\mathcal{T}=\{\tau_i\}_{i=1}^N$, where each $\tau_i = \{(s_t,a_t)\}_{t=1}^{T_i}$
is a sequence of state–action pairs terminating at step $T_i$. 
We define a refinement operator $\mathcal{R}$ that maps the raw set $\mathcal{T}$ into a refined set $\mathcal{T}^*$:
\begin{equation}
\mathcal{T}^* = \mathcal{R}(\mathcal{T}).
\label{eq:refinement}
\end{equation}
For each $\tau_i$, the refinement identifies the shortest feasible subsequence $\tau_i^*$ that still leads to the same correct final answer:
\begin{equation}
\tau_i^* = \arg\min_{\tau \in \mathcal{F}_i} |\tau|,
\label{eq:argmin-refine}
\end{equation}
where $\mathcal{F}_i$ is the set of all feasible answer-consistent trajectories equivalent to $\tau_i$. Let $s_{T}(\tau)$ denotes the final state and $y(\tau)$ denotes the answer.
\begin{equation}
\mathcal{F}_i = \big\{ \tau \mid s_{T}(\tau) = s_{T}(\tau_i) \land y(\tau) = y(\tau_i) \big\}.
\end{equation}
Thus, the refined dataset $\mathcal{T}^*$ retains trajectories that are semantically equivalent to the originals but stripped of redundant exploration steps. 
Each refined trajectory $\tau_i^*$ is then decomposed into step-level supervision signals by attaching a rule-based stepwise reward $\ell_t \in [0,1]$ to each action $a_t$, indicating its correctness within the reasoning trajectory. 
Formally, the RL training dataset is constructed as $\mathcal D_{\text{RL}} = \big\{(s_t, a_t, \ell_t)\big\}_{\tau^*,\,t}$.
This ensures that RL receives concise stepwise supervision signals, improving both stability and efficiency of training.

\subsection{Training Stage}

To enhance the model's graph comprehension and reasoning capabilities, we adopt a two-stage fine-tuning approach. The first stage uses SFT with synthesized data to establish foundational abilities. The second stage employs GRPO with trajectory refinement, leveraging RL's proven effectiveness in enhancing reasoning and exploration efficiency~\citep{yue2025does}.

\subsubsection{Stage I: Supervised fine-tuning}
For each step $t$, let $s_t=(q,\mathcal G^p_t,\mathcal G^{sub}_t,h_t)$ be the serialized state,
and let $y_t=(y_t^1,\dots,y_t^{L_t})$ be the target token sequence that concatenates
the natural-language thought process and the action specification.
Denote by $\mathcal I(s_t)$ the textual serialization of the state.
The training loss of SFT is defined as
\begin{equation}
\small 
\mathcal{L}_{\mathrm{SFT}}(\theta)
= -\mathbb{E}_{\substack{(s_t,y_t)\\ \sim \mathcal D_{\text{SFT}}}}
\Bigg[
\sum_{l=1}^{L_t}
\log \pi_{\theta}\big(y_t^{\,l} \mid \mathcal I(s_t),\,y_t^{<l}\big)
\Bigg].
\end{equation}
where $\pi_{\theta}(y_t^{\,l}\mid \mathcal I(s_t),y_t^{<l})$ denotes the probability
assigned by the model to the $l$-th token given the serialized state and the previously
generated tokens.

\subsubsection{Stage II: RL with stepwise rewards}
For reward
design, existing approaches predominantly rely on outcome-based reward signals, which have demonstrated remarkable
effectiveness in domains such as mathematical reasoning
and code generation. However, prior studies~\citep{wang2025spa,choudhury2025process,deng2024novice} have shown that
in relatively complex scenarios such as graph retrieval, conventional outcome-based reward signals tend to be overly
sparse. This sparsity hampers effective credit assignment to
early-stage actions, ultimately resulting in inefficient learning over long action chains. This observation motivates our adoption of process-level rewards in the training process, where the reward signal is determined by the contribution of each current action to the golden subgraphs.

Specifically, each step $t$ is associated with a process-level rule-based reward $\ell_t$ that provides
graded feedback according to the quality of the predicted action:
\begin{equation}
\label{eq:reward}
\ell_t =
\begin{cases}
0, & \text{if } \mathrm{Inv}(a_t), \\
c_1, & \text{if } \mathrm{Fmt}(a_t)=1 \land \mathrm{Corr}(a_t)=0, \\
c_2, & \text{if } \mathrm{Part}(a_t)=1, \\
1.0, & \text{if } a_t = a_t^{*}.
\end{cases}
\end{equation}
\noindent
Here, $\mathrm{Inv}(\cdot)$, $\mathrm{Fmt}(\cdot)$, $\mathrm{Corr}(\cdot)$, and $\mathrm{Part}(\cdot)$ denote the deterministic functions for checking validity, format correctness, action correctness, and partial correctness, respectively. $c_1$ and $c_2$ are dataset-specific hyperparameters. Consequently, we adopt the reward function shown in Eq.~\ref{eq:reward} to train our model using the GRPO method.

\subsection{Interactive Retriever}

At inference time, Graph-S$^3$ interacts with the textual graph through the defined action space. Unlike single-pass retrieval methods that return large subgraphs, our approach performs stepwise exploration by balancing EXPLORE and CHOOSE actions, terminating with FINISH when sufficient evidence is gathered. This interactive process enables precise control over retrieval depth while minimizing redundancy, producing concise subgraphs for reasoning. The detailed prompts designed for the interactive graph retriever are provided in Appendix~\ref{prompt}. An example of interactive reasoning process of Graph-$S^3$ can be found in Appendix~\ref{ReasoningProcess}.

\section{Experiments}

\begin{table*}[t!]
\renewcommand{\arraystretch}{1.15}
\centering
\resizebox{\textwidth}{!}{%
\begin{tabular}{r|rr|rr|rl|rl|rl}
\hline
\multirow{2}{*}{\textbf{Retriever + Generator}} & \multicolumn{2}{c|}{\textbf{WebQSP}} & \multicolumn{2}{c|}{\textbf{CWQ}} & \multicolumn{2}{c|}{\textbf{MetaQA 1-hop}} & \multicolumn{2}{c|}{\textbf{MetaQA 2-hop}} & \multicolumn{2}{c}{\textbf{MetaQA 3-hop}} \\
 & \textbf{Acc} & $\mathbf{F_1}$ & \textbf{Acc} & $\mathbf{F_1}$ & \textbf{Acc} & $\mathbf{F_1}$ & \textbf{Acc} & $\mathbf{F_1}$ & \textbf{Acc} & $\mathbf{F_1}$ \\ \hline
No graph + Qwen3-8B & 5.16 & 8.11 & 6.26 & 7.35 & 2.00 & 2.88 & 0.07 & 0.95 & 0.20 & 1.19 \\
No retriever + Qwen3-8B & 0.25 & 1.83 & 0.37 & 1.29 & 0.00 & 0.29 & 0.00 & 0.49 & 0.00 & 1.25 \\
RAG/1hop + Qwen3-8B & \underline{27.89} & \underline{38.57} & 12.55 & 16.18 & \underline{75.93} & \underline{86.36} & 0.77 & 1.74 & \underline{4.13} & \underline{11.99} \\
RAG/2hop + Qwen3-8B & 14.07 & 24.47 & 7.00 & 10.55 & 42.03 & 55.59 & \underline{10.07} & \underline{21.65} & 2.60 & 9.42 \\
RAG/3hop + Qwen3-8B & 1.54 & 7.94 & 0.99 & 3.12 & 0.37 & 3.03 & 0.13 & 2.21 & 0.13 & 3.45 \\
ToG + Qwen3-8B & 6.14 & 9.79 & 7.01 & 9.61 & 1.37 & 1.75 & 0.00 & 0.00 & 0.00 & 0.20 \\
LightRAG + Qwen3-8B & 18.39 & 31.67 & \underline{16.20} & \underline{23.09} & 1.13 & 1.76 & 0.00 & 0.19 & 0.07 & 0.40 \\
G-retriever + Qwen3-8B & 25.74 & 35.45 & 15.38 & 18.62 & 0.63 & 1.60 & 0.10 & 0.77 & 0.03 & 1.87 \\
Graph-$S^3$ + Qwen3-8B & \textbf{36.24} & \textbf{47.88} & \textbf{17.87} & \textbf{23.29} & \textbf{81.50} & \textbf{90.22} & \textbf{53.73} & \textbf{65.60} & \textbf{12.73} & \textbf{29.49} \\ \hline
No graph + Llama3.1-8B & 8.97 & 15.69 & 9.40 & 11.58 & 12.20 & 17.60 & 1.27 & 7.77 & 1.23 & 8.86 \\
No retriever + Llama3.1-8B & 0.18 & 1.97 & 0.14 & 1.29 & 0.00 & 0.58 & 0.00 & 1.11 & 0.00 & 2.94 \\
RAG/1hop + Llama3.1-8B & \underline{24.82} & 35.28 & 13.85 & 17.26 & \underline{60.17} & \underline{70.84} & 2.40 & 5.98 & \underline{4.03} & \underline{15.46} \\
RAG/2hop + Llama3.1-8B & 11.06 & 22.94 & 6.29 & 10.68 & 29.07 & 42.47 & \underline{4.50} & \underline{15.06} & 1.80 & 11.30 \\
RAG/3hop + Llama3.1-8B & 1.04 & 6.67 & 0.65 & 3.43 & 0.33 & 3.67 & 0.17 & 3.37 & 0.07 & 5.76 \\
ToG + Llama3.1-8B & 8.85 & 14.28 & 8.42 & 12.33 & 12.40 & 15.88 & 0.00 & 0.63 & 1.43 & 6.10 \\
LightRAG + Llama3.1-8B & 15.85 & \underline{36.66} & 8.33 & 15.01 & 13.13 & 21.47 & 0.93 & 4.33 & 1.00 & 6.38 \\
G-retriever + Llama3.1-8B & 22.67 & 32.26 & \underline{13.91} & \underline{17.47} & 0.67 & 1.56 & 0.10 & 0.83 & 0.10 & 1.77 \\
Graph-$S^3$ + Llama3.1-8B & \textbf{32.31} & \textbf{43.26} & \textbf{17.11} & \textbf{21.17} & \textbf{67.50} & \textbf{76.56} & \textbf{40.17} & \textbf{55.49} & \textbf{10.73} & \textbf{29.55} \\ \hline
No graph + Finetuned-8B & 9.21 & 14.88 & 10.17 & 12.31 & 1.63 & 2.49 & 0.43 & 2.64 & 0.33 & 4.60 \\
No retriever + Finetuned-8B & 0.37 & 2.26 & 0.68 & 1.76 & 0.00 & 0.29 & 0.00 & 0.34 & 0.00 & 1.05 \\
RAG/1hop + Finetuned-8B & 28.87 & 41.48 & 19.85 & 26.01 & \underline{59.50} & \underline{69.54} & 1.83 & 7.44 & \underline{3.30} & \underline{18.61} \\
RAG/2hop + Finetuned-8B & 14.93 & 27.76 & 8.89 & 14.56 & 35.83 & 52.03 & \underline{7.70} & \underline{21.21} & 3.07 & 14.04 \\
RAG/3hop + Finetuned-8B & 1.54 & 7.81 & 0.93 & 4.36 & 0.57 & 2.99 & 0.43 & 2.31 & 0.13 & 4.23 \\
ToG + Finetuned-8B & 5.04 & 9.43 & 7.54 & 9.79 & 2.23 & 3.44 & 0.00 & 0.12 & 0.10 & 2.80 \\
LightRAG + Finetuned-8B & 17.38 & 32.59 & 13.85 & 19.96 & 13.77 & 20.60 & 0.97 & 3.71 & 0.53 & 4.28 \\
G-retriever + Finetuned-8B & \underline{30.34} & \underline{43.49} & \underline{22.68} & \underline{28.38} & 8.93 & 11.51 & 2.33 & 4.80 & 0.40 & 3.31 \\
Graph-$S^3$ + Finetuned-8B & \textbf{44.29} & \textbf{58.45} & \textbf{23.62} & \textbf{30.73} & \textbf{82.77} & \textbf{92.04} & \textbf{63.17} & \textbf{76.18} & \textbf{14.70} & \textbf{36.29} \\ \hline
\end{tabular}%
}
\caption{Overall results on graph-based QA benchmarks. The best results are highlighted in bold and the second performance results are indicated by an underscore.}
\label{table:main_result}
\end{table*}

\subsection{Experimental Setup}
\textbf{Datasets.} We evaluate Graph-$S^3$ on three widely used textual graph QA benchmarks. 
WebQSP~\citep{yih2015semantic} consists of real-world questions annotated with SPARQL queries against Freebase, 
primarily involving one- or two-hop reasoning. 
CWQ~\citep{talmor2018web} extends WebQSP with more complex multi-hop questions, 
posing a greater challenge for long reasoning chains. 
MetaQA~\citep{zhang2018variational} is a movie-domain benchmark containing 135k triples and 43k entities, 
designed to evaluate multi-hop reasoning in a closed domain. 
Following prior work~\citep{chen2024kg}, we report accuracy (Acc) and $\mathbf{F_1}$ score 
as evaluation metrics.

\textbf{Baselines.} To validate the effectiveness of our approach, we compare with several representative graph retrieval methods. 
We additionally evaluate the model's inherent graph understanding capability through two configurations:  
(1) the \emph{no graph} setting, where the model processes the query without any graph input, and  
(2) the \emph{no retriever} setting, where the model receives the entire graph structure directly as input.  

For traditional RAG, we implemented a multi-hop method where the model retrieves the most relevant graph nodes for the current query and then performs a k-hop expansion to collect information for answer generation.
We further compare with representative graph retrievers, including Think-on-Graph (ToG)~\citep{sun2023think}, LightRAG~\citep{guo2024lightrag}, G-Retriever~\citep{he2024g}.

Implementation Details are provided in the Appendix~\ref{sec:details}.


\subsection{Main Results}

The results of WebQSP, CWQ, and MetaQA are summarized in Table~\ref{table:main_result}. In general, our framework achieves the best performance among all compared methods, demonstrating the effectiveness of combining two-stage training with interactive retrieval1. We break down the analysis into several key observations.

\textbf{Superiority in Complex Multi-hop Reasoning.} Graph-$S^3$ exhibits a widening performance gap over baselines as reasoning complexity increases. On the MetaQA benchmark, while competitive on 1-hop questions, our method dominates on multi-hop tasks. Specifically, on MetaQA 3-hop, Graph-$S^3$ significantly outperforming the next best agentic baseline and far surpassing traditional $k$-hop RAG methods which struggle to maintain coherence over long chains. 

\textbf{Effectiveness Against Training-Free Baselines.} Against sophisticated training-free baselines like ToG and LightRAG, our model delivers substantial gains across all datasets. This highlights that while LLM-based planning (ToG) or structural awareness (LightRAG) is beneficial, it is insufficient without the dedicated fine-tuning and reinforcement learning provided by our pipeline to align the retriever with graph-specific reasoning patterns.

\textbf{Advantage Over Trained Retrievers.} Compared to supervised retrievers like G-Retriever, Graph-$S^3$ consistently achieves higher $F_1$ scores, indicating better precision in answer generation. This improvement suggests that our interactive pruning mechanism effectively balances recall and precision—retrieving enough context to answer correctly without overwhelming the generator with noise, a common pitfall for G-Retriever which retrieves larger, noisier subgraphs.

\textbf{Robustness Across Backbones.} The performance gains are consistent across different LLM backbones and fine-tuning strategies. Notably, even when using a fine-tuned backbone, Graph-$S^3$ maintains its lead, achieving the highest scores on all benchmarks. This demonstrates that our method's benefits are orthogonal to the underlying model's capability and can serve as a universal enhancement for graph-based QA systems.

To gain deeper insights into the quantitative improvements, we analyzed specific reasoning examples to verify the effectiveness of our interactive pruning and stepwise supervision. A detailed presentation of these case studies and the corresponding reasoning processes of Graph-$S^3$ can be found in Appendix~\ref{CaseStudy}.

\subsection{In-depth Analysis}
\begin{table*}[ht]
\renewcommand{\arraystretch}{1.1}
\centering
\resizebox{\textwidth}{!}{%
\begin{tabular}{c|cccccccccc}
\hline
\multirow{4}{*}{\textbf{Methods of Ablation}} & \multicolumn{10}{c}{\textbf{Dataset}} \\ \cline{2-11} 
 & \multicolumn{2}{c|}{\multirow{2}{*}{\textbf{WebQSP}}} & \multicolumn{2}{c|}{\multirow{2}{*}{\textbf{CWQ}}} & \multicolumn{6}{c}{\textbf{MetaQA}} \\ \cline{6-11} 
 & \multicolumn{2}{c|}{} & \multicolumn{2}{c|}{} & \multicolumn{2}{c|}{\textbf{1hop}} & \multicolumn{2}{c|}{\textbf{2hop}} & \multicolumn{2}{c}{\textbf{3hop}} \\ \cline{2-11} 
 & \textbf{Acc} & \multicolumn{1}{c|}{$\mathbf{F_1}$} & \textbf{Acc} & \multicolumn{1}{c|}{$\mathbf{F_1}$} & \textbf{Acc} & \multicolumn{1}{c|}{$\mathbf{F_1}$} & \textbf{Acc} & \multicolumn{1}{c|}{$\mathbf{F_1}$} & \textbf{Acc} & $\mathbf{F_1}$ \\ \hline
\textbf{Graph-$S^3$} & \textbf{44.29} & \multicolumn{1}{c|}{\textbf{58.45}} & \textbf{23.62} & \multicolumn{1}{c|}{\textbf{30.73}} & \textbf{82.77} & \multicolumn{1}{c|}{\textbf{92.04}} & \textbf{63.17} & \multicolumn{1}{c|}{\textbf{76.18}} & \textbf{14.70} & \textbf{36.29} \\
\textbf{w/o SFT} & \underline{31.64} & \multicolumn{1}{c|}{44.41} & 7.74 & \multicolumn{1}{c|}{8.77} & \underline{81.27} & \multicolumn{1}{c|}{\underline{89.38}} & \underline{46.30} & \multicolumn{1}{c|}{\underline{54.12}} & 2.07 & 4.98 \\
\textbf{w/o RL} & 41.77 & \multicolumn{1}{c|}{\underline{53.02}} & 13.39 & \multicolumn{1}{c|}{15.97} & 71.97 & \multicolumn{1}{c|}{80.09} & 35.93 & \multicolumn{1}{c|}{45.25} & \underline{5.73} & 11.46 \\
\textbf{w/o interactive} & 28.87 & \multicolumn{1}{c|}{41.48} & \underline{19.85} & \multicolumn{1}{c|}{\underline{26.01}} & 
59.50 & \multicolumn{1}{c|}{69.54} & 1.83 & \multicolumn{1}{c|}{7.44} & 3.30 & \underline{18.61} \\
\textbf{w/o traj. refine} & 16.46 & \multicolumn{1}{c|}{19.24} & 4.12 & \multicolumn{1}{c|}{4.87} & 39.47 & \multicolumn{1}{c|}{41.06} & 4.01 & \multicolumn{1}{c|}{6.10} & 1.34 & 1.80 \\ \hline
\end{tabular}%
}
\caption{Results of ablation studies.}
\label{table:ab}
\end{table*}

\subsubsection{Ablation Study}
Our framework consists of four key components: supervised fine-tuning (SFT), reinforcement learning (RL) with stepwise rewards, interactive retrieval at inference time, and trajectory refinement during data synthesis. 
To assess the contribution of each component, we remove one module at a time and evaluate the resulting performance degradation. 
The results are reported in Table~\ref{table:ab}.

\textbf{Ablation of SFT.} 
Removing the SFT stage leads to a clear drop in Accuracy and $\mathbf{F_1}$ across all benchmarks. 
This confirms that SFT provides the retriever with essential navigation ability, compensating for the lack of graph-specific training during upstream pretraining and establishing a stable foundation for subsequent RL optimization.  

\textbf{Ablation of RL.}
Eliminating the RL stage results in consistent performance degradation, with particularly large declines on CWQ and MetaQA, which require longer reasoning chains. 
This demonstrates that RL with stepwise rewards substantially strengthens the retriever’s reasoning capability, especially on complex multi-hop tasks.  

\textbf{Ablation of interactive inference.}
Disabling interactive retrieval causes significant performance drops on 2-hop and 3-hop questions, where results approach those of conventional $k$-hop RAG. 
This shows that interactive retrieval is crucial for adaptively controlling retrieval depth, effectively filtering redundant neighbors while preserving critical relations.  

\textbf{Ablation of trajectory refinement.}
Removing trajectory refinement during data synthesis leads to the largest degradation among all ablations. The results indicate that without refinement, synthetic trajectories contain redundant detours, which produce noisy reward signals and undermine the stability of RL optimization.

\begin{table*}[ht]
\renewcommand{\arraystretch}{1.1}
\centering
\resizebox{\textwidth}{!}{%
\begin{tabular}{l|cccccccccc} 
\hline
\multirow{4}{*}{\textbf{Retriever Train Method}} & \multicolumn{10}{c}{\textbf{Dataset}} \\ \cline{2-11} 
 & \multicolumn{2}{c|}{\multirow{2}{*}{\textbf{WebQSP}}} & \multicolumn{2}{c|}{\multirow{2}{*}{\textbf{CWQ}}} & \multicolumn{6}{c}{\textbf{MetaQA}} \\ \cline{6-11} 
 & \multicolumn{2}{c|}{} & \multicolumn{2}{c|}{} & \multicolumn{2}{c|}{\textbf{1hop}} & \multicolumn{2}{c|}{\textbf{2hop}} & \multicolumn{2}{c}{\textbf{3hop}} \\ \cline{2-11} 
 & \textbf{Acc} & \multicolumn{1}{c|}{$\mathbf{F_1}$} & \textbf{Acc} & \multicolumn{1}{c|}{$\mathbf{F_1}$} & \textbf{Acc} & \multicolumn{1}{c|}{$\mathbf{F_1}$} & \textbf{Acc} & \multicolumn{1}{c|}{$\mathbf{F_1}$} & \textbf{Acc} & $\mathbf{F_1}$ \\ \hline
\textbf{w/o step supervision} & 41.83 & \multicolumn{1}{c|}{53.87} & 13.47 & \multicolumn{1}{c|}{16.40} & 72.63 & \multicolumn{1}{c|}{81.12} & 34.97 & \multicolumn{1}{c|}{45.14} & 6.43 & 11.34 \\
\textbf{Graph-$S^3$} & \textbf{44.29} & \multicolumn{1}{c|}{\textbf{58.45}} & \textbf{23.62} & \multicolumn{1}{c|}{\textbf{30.73}} & \textbf{82.77} & \multicolumn{1}{c|}{\textbf{92.04}} & \textbf{63.17} & \multicolumn{1}{c|}{\textbf{76.18}} & \textbf{14.70} & \textbf{36.29} \\ \hline
\end{tabular}%
}
\caption{Performance comparison of process-level rewards and outcome-based rewards training methods.}
\label{table:prm}
\end{table*}

\subsubsection{Effectiveness of Stepwise Supervision}

To further validate the effectiveness of our proposed stepwise supervision, we conducted an ablation study. Specifically, starting from the SFT-trained model, we ablated the stepwise reward signals and modified the setup to rely solely on outcome-based rewards. The results of this ablation study are shown in Table~\ref{table:prm}. Experimental results indicate that without stepwise rewards, model performance experiences a significant decline across all benchmarks, particularly on CWQ and MetaQA which involve longer reasoning chains. This confirms that fine-grained stepwise supervision enables more stable optimization and better generalization on complex multi-hop reasoning tasks.

\begin{figure}[t]
\centering
  \includegraphics[width=\columnwidth]{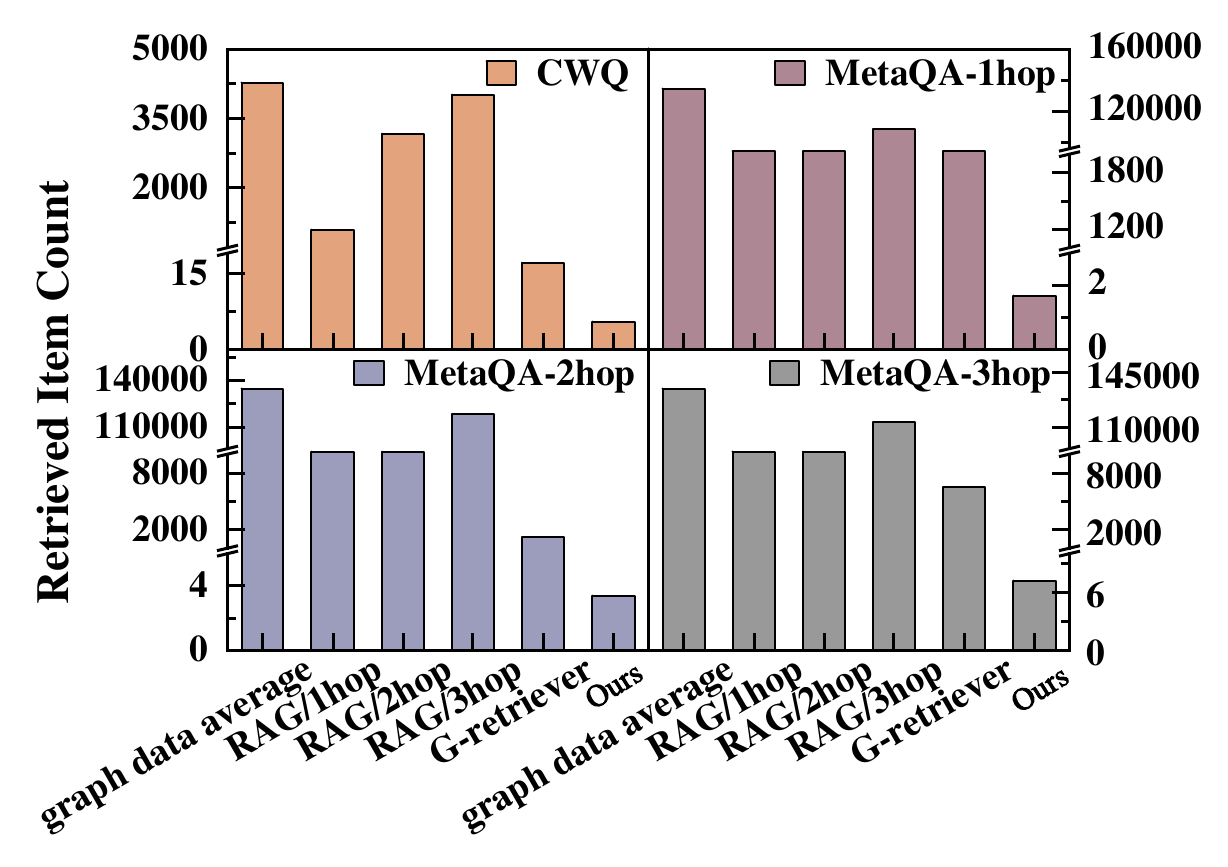}
  \caption{Number of retrieved graph triples in Graph-$S^3$ and baselines on correct answers.}
  \label{fig:num_retrieved}
\end{figure}

\subsubsection{Effective Information Quantification Analysis}
To evaluate the efficiency of Graph-$S^3$ in retrieving concise yet effective information, we compare it with baseline approaches by measuring the number of triples required to produce correct answers (see Figure~\ref{fig:num_retrieved}). Unlike traditional methods that often retrieve large amounts of redundant information, our approach significantly reduces retrieval size while maintaining higher accuracy. In particular, our experiments show that Graph-$S^3$ requires only \textbf{11.44\%} of the triples retrieved by G-Retriever on average, yet still achieves superior accuracy. These results highlight the framework’s ability to balance search depth with precision, thereby reducing redundancy.

\section{Conclusion}
We investigated the limitations of existing retrieval-augmented generation methods on textual graphs, 
highlighting their reliance on outcome-based supervision and their tendency to produce redundant or incomplete subgraphs.  
To overcome these challenges, we proposed a framework that integrates three key innovations:  

(1) A pipeline for high-quality, stepwise-supervised data synthesis;
(2) Two-stage training (SFT then RL) with process-level rewards;
(3) Fine-grained, interactive retrieval over textual graphs.
Extensive experiments on WebQSP, CWQ, and MetaQA demonstrate that our approach consistently improves both accuracy and $F_1$, 
validating the effectiveness of synthetic stepwise supervision and the proposed training strategy for enhancing interactive graph retrieval. These findings underscore the potential of fine-grained process supervision in bridging the gap between LLMs and structured knowledge, paving the way for more reliable and interpretable agentic reasoning systems.

\section{Limitation}
While this work has focused on validating Graph-S³ on standard textual graph benchmarks, future research could explore extending the interactive retrieval framework to diverse problem scenarios. Additionally, with the rapid advancement of agent technology, the current framework can be further enhanced by designing a broader spectrum of actions and integrating external tools to tackle increasingly complex reasoning tasks.

\section{Acknowledgments}
This work was supported in part by the National Natural Science Foundation of China under Grant No. 62272261, Wuxi Research Institute of Applied Technologies, Tsinghua University under Grant 20242001120, Tsinghua University (AIR)–AsiaInfo Technologies (China) Inc. Joint Research Center and a grant from GDS Holdings Limited.

\bibliography{custom}

@inproceedings{yih2016value,
  title={The value of semantic parse labeling for knowledge base question answering},
  author={Yih, Wen-tau and Richardson, Matthew and Meek, Christopher and Chang, Ming-Wei and Suh, Jina},
  booktitle={Proceedings of the 54th Annual Meeting of the Association for Computational Linguistics (Volume 2: Short Papers)},
  pages={201--206},
  year={2016}
}

@article{chang2024survey,
  title={A survey on evaluation of large language models},
  author={Chang, Yupeng and Wang, Xu and Wang, Jindong and Wu, Yuan and Yang, Linyi and Zhu, Kaijie and Chen, Hao and Yi, Xiaoyuan and Wang, Cunxiang and Wang, Yidong and others},
  journal={ACM transactions on intelligent systems and technology},
  volume={15},
  number={3},
  pages={1--45},
  year={2024},
  publisher={ACM New York, NY}
}

@article{lewis2020retrieval,
  title={Retrieval-augmented generation for knowledge-intensive nlp tasks},
  author={Lewis, Patrick and Perez, Ethan and Piktus, Aleksandra and Petroni, Fabio and Karpukhin, Vladimir and Goyal, Naman and K{\"u}ttler, Heinrich and Lewis, Mike and Yih, Wen-tau and Rockt{\"a}schel, Tim and others},
  journal={Advances in neural information processing systems},
  volume={33},
  pages={9459--9474},
  year={2020}
}

@article{peng2024graph,
  title={Graph retrieval-augmented generation: A survey},
  author={Peng, Boci and Zhu, Yun and Liu, Yongchao and Bo, Xiaohe and Shi, Haizhou and Hong, Chuntao and Zhang, Yan and Tang, Siliang},
  journal={arXiv preprint arXiv:2408.08921},
  year={2024}
}

@inproceedings{procko2024graph,
  title={Graph retrieval-augmented generation for large language models: A survey},
  author={Procko, Tyler Thomas and Ochoa, Omar},
  booktitle={2024 Conference on AI, Science, Engineering, and Technology (AIxSET)},
  pages={166--169},
  year={2024},
  organization={IEEE}
}

@article{zhang2025survey,
  title={A Survey of Graph Retrieval-Augmented Generation for Customized Large Language Models},
  author={Zhang, Qinggang and Chen, Shengyuan and Bei, Yuanchen and Yuan, Zheng and Zhou, Huachi and Hong, Zijin and Dong, Junnan and Chen, Hao and Chang, Yi and Huang, Xiao},
  journal={arXiv preprint arXiv:2501.13958},
  year={2025}
}

@article{jin2024large,
  title={Large language models on graphs: A comprehensive survey},
  author={Jin, Bowen and Liu, Gang and Han, Chi and Jiang, Meng and Ji, Heng and Han, Jiawei},
  journal={IEEE Transactions on Knowledge and Data Engineering},
  year={2024},
  publisher={IEEE}
}

@article{chai2023graphllm,
  title={Graphllm: Boosting graph reasoning ability of large language model},
  author={Chai, Ziwei and Zhang, Tianjie and Wu, Liang and Han, Kaiqiao and Hu, Xiaohai and Huang, Xuanwen and Yang, Yang},
  journal={arXiv preprint arXiv:2310.05845},
  year={2023}
}

@article{chen2024kg,
  title={Kg-retriever: Efficient knowledge indexing for retrieval-augmented large language models},
  author={Chen, Weijie and Bai, Ting and Su, Jinbo and Luan, Jian and Liu, Wei and Shi, Chuan},
  journal={arXiv preprint arXiv:2412.05547},
  year={2024}
}

@article{guo2024lightrag,
  title={Lightrag: Simple and fast retrieval-augmented generation},
  author={Guo, Zirui and Xia, Lianghao and Yu, Yanhua and Ao, Tu and Huang, Chao},
  journal={arXiv preprint arXiv:2410.05779},
  year={2024}
}

@article{he2024g,
  title={G-retriever: Retrieval-augmented generation for textual graph understanding and question answering},
  author={He, Xiaoxin and Tian, Yijun and Sun, Yifei and Chawla, Nitesh and Laurent, Thomas and LeCun, Yann and Bresson, Xavier and Hooi, Bryan},
  journal={Advances in Neural Information Processing Systems},
  volume={37},
  pages={132876--132907},
  year={2024}
}

@article{jiang2024kg,
  title={Kg-agent: An efficient autonomous agent framework for complex reasoning over knowledge graph},
  author={Jiang, Jinhao and Zhou, Kun and Zhao, Wayne Xin and Song, Yang and Zhu, Chen and Zhu, Hengshu and Wen, Ji-Rong},
  journal={arXiv preprint arXiv:2402.11163},
  year={2024}
}

@article{yang2024graphagent,
  title={GraphAgent: Agentic Graph Language Assistant},
  author={Yang, Yuhao and Tang, Jiabin and Xia, Lianghao and Zou, Xingchen and Liang, Yuxuan and Huang, Chao},
  journal={arXiv preprint arXiv:2412.17029},
  year={2024}
}

@article{chen2020review,
  title={A review: Knowledge reasoning over knowledge graph},
  author={Chen, Xiaojun and Jia, Shengbin and Xiang, Yang},
  journal={Expert systems with applications},
  volume={141},
  pages={112948},
  year={2020},
  publisher={Elsevier}
}

@article{hogan2021knowledge,
  title={Knowledge graphs},
  author={Hogan, Aidan and Blomqvist, Eva and Cochez, Michael and d’Amato, Claudia and Melo, Gerard De and Gutierrez, Claudio and Kirrane, Sabrina and Gayo, Jos{\'e} Emilio Labra and Navigli, Roberto and Neumaier, Sebastian and others},
  journal={ACM Computing Surveys (Csur)},
  volume={54},
  number={4},
  pages={1--37},
  year={2021},
  publisher={ACM New York, NY, USA}
}

@inproceedings{zou2020survey,
  title={A survey on application of knowledge graph},
  author={Zou, Xiaohan},
  booktitle={Journal of Physics: Conference Series},
  volume={1487},
  number={1},
  pages={012016},
  year={2020},
  organization={IOP Publishing}
}

@inproceedings{ji2024retrieval,
  title={Retrieval and reasoning on KGs: Integrate knowledge graphs into large language models for complex question answering},
  author={Ji, Yixin and Wu, Kaixin and Li, Juntao and Chen, Wei and Zhong, Mingjie and Jia, Xu and Zhang, Min},
  booktitle={Findings of the Association for Computational Linguistics: EMNLP 2024},
  pages={7598--7610},
  year={2024}
}

@article{pahilajani2024grs,
  title={GRS-QA--Graph Reasoning-Structured Question Answering Dataset},
  author={Pahilajani, Anish and Trivedi, Devasha and Shuai, Jincen and Yone, Khin S and Jain, Samyak Rajesh and Park, Namyong and Rossi, Ryan A and Ahmed, Nesreen K and Dernoncourt, Franck and Wang, Yu},
  journal={arXiv preprint arXiv:2411.00369},
  year={2024}
}

@article{han2024retrieval,
  title={Retrieval-augmented generation with graphs (graphrag)},
  author={Han, Haoyu and Wang, Yu and Shomer, Harry and Guo, Kai and Ding, Jiayuan and Lei, Yongjia and Halappanavar, Mahantesh and Rossi, Ryan A and Mukherjee, Subhabrata and Tang, Xianfeng and others},
  journal={arXiv preprint arXiv:2501.00309},
  year={2024}
}

@article{zhu2025graph,
  title={Graph-based Approaches and Functionalities in Retrieval-Augmented Generation: A Comprehensive Survey},
  author={Zhu, Zulun and Huang, Tiancheng and Wang, Kai and Ye, Junda and Chen, Xinghe and Luo, Siqiang},
  journal={arXiv preprint arXiv:2504.10499},
  year={2025}
}

@article{jaech2024openai,
  title={Openai o1 system card},
  author={Jaech, Aaron and Kalai, Adam and Lerer, Adam and Richardson, Adam and El-Kishky, Ahmed and Low, Aiden and Helyar, Alec and Madry, Aleksander and Beutel, Alex and Carney, Alex and others},
  journal={arXiv preprint arXiv:2412.16720},
  year={2024}
}

@article{el2025competitive,
  title={Competitive programming with large reasoning models},
  author={El-Kishky, Ahmed and Wei, Alexander and Saraiva, Andre and Minaiev, Borys and Selsam, Daniel and Dohan, David and Song, Francis and Lightman, Hunter and Clavera, Ignasi and Pachocki, Jakub and others},
  journal={arXiv preprint arXiv:2502.06807},
  year={2025}
}

@article{guo2025deepseek,
  title={Deepseek-r1: Incentivizing reasoning capability in llms via reinforcement learning},
  author={Guo, Daya and Yang, Dejian and Zhang, Haowei and Song, Junxiao and Zhang, Ruoyu and Xu, Runxin and Zhu, Qihao and Ma, Shirong and Wang, Peiyi and Bi, Xiao and others},
  journal={arXiv preprint arXiv:2501.12948},
  year={2025}
}

@article{LIU2025111625,
title = {Knowledge graph reasoning: Mainstream methods, applications and prospects},
journal = {Engineering Applications of Artificial Intelligence},
volume = {159},
pages = {111625},
year = {2025},
issn = {0952-1976},
doi = {https://doi.org/10.1016/j.engappai.2025.111625},
url = {https://www.sciencedirect.com/science/article/pii/S0952197625016276},
author = {Han Liu and Shaojie Yang and Guokai Shi and Zongcheng Miao}
}

@article{yao2025learning,
  title={Learning Efficient and Generalizable Graph Retriever for Knowledge-Graph Question Answering},
  author={Yao, Tianjun and Li, Haoxuan and Shen, Zhiqiang and Li, Pan and Liu, Tongliang and Zhang, Kun},
  journal={arXiv preprint arXiv:2506.09645},
  year={2025}
}

@article{guo2024deepseek,
  title={DeepSeek-Coder: When the Large Language Model Meets Programming--The Rise of Code Intelligence},
  author={Guo, Daya and Zhu, Qihao and Yang, Dejian and Xie, Zhenda and Dong, Kai and Zhang, Wentao and Chen, Guanting and Bi, Xiao and Wu, Yu and Li, YK and others},
  journal={arXiv preprint arXiv:2401.14196},
  year={2024}
}

@article{das2017go,
  title={Go for a walk and arrive at the answer: Reasoning over paths in knowledge bases using reinforcement learning},
  author={Das, Rajarshi and Dhuliawala, Shehzaad and Zaheer, Manzil and Vilnis, Luke and Durugkar, Ishan and Krishnamurthy, Akshay and Smola, Alex and McCallum, Andrew},
  journal={arXiv preprint arXiv:1711.05851},
  year={2017}
}

@article{cui2025process,
  title={Process reinforcement through implicit rewards},
  author={Cui, Ganqu and Yuan, Lifan and Wang, Zefan and Wang, Hanbin and Li, Wendi and He, Bingxiang and Fan, Yuchen and Yu, Tianyu and Xu, Qixin and Chen, Weize and others},
  journal={arXiv preprint arXiv:2502.01456},
  year={2025}
}

@article{zhang2025r1,
  title={R1-reward: Training multimodal reward model through stable reinforcement learning},
  author={Zhang, Yi-Fan and Lu, Xingyu and Hu, Xiao and Fu, Chaoyou and Wen, Bin and Zhang, Tianke and Liu, Changyi and Jiang, Kaiyu and Chen, Kaibing and Tang, Kaiyu and others},
  journal={arXiv preprint arXiv:2505.02835},
  year={2025}
}

@article{choubey2024distill,
  title={Distill-SynthKG: Distilling Knowledge Graph Synthesis Workflow for Improved Coverage and Efficiency},
  author={Choubey, Prafulla Kumar and Su, Xin and Luo, Man and Peng, Xiangyu and Xiong, Caiming and Le, Tiep and Rosenman, Shachar and Lal, Vasudev and Mui, Phil and Ho, Ricky and others},
  journal={arXiv preprint arXiv:2410.16597},
  year={2024}
}

@misc{OpenAI_GPT4o_System_Card_2024,
  title        = {GPT-4o System Card},
  author       = {OpenAI},
  year         = {2024},
  howpublished = {OpenAI Technical documentation},
  note         = {Available online: https://openai.com/index/gpt-4o-system-card/}
}

@article{sun2023think,
  title={Think-on-graph: Deep and responsible reasoning of large language model on knowledge graph},
  author={Sun, Jiashuo and Xu, Chengjin and Tang, Lumingyuan and Wang, Saizhuo and Lin, Chen and Gong, Yeyun and Ni, Lionel M and Shum, Heung-Yeung and Guo, Jian},
  journal={arXiv preprint arXiv:2307.07697},
  year={2023}
}

@inproceedings{yih2015semantic,
  title={Semantic parsing via staged query graph generation: Question answering with knowledge base},
  author={Yih, Scott Wen-tau and Chang, Ming-Wei and He, Xiaodong and Gao, Jianfeng},
  booktitle={Proceedings of the Joint Conference of the 53rd Annual Meeting of the ACL and the 7th International Joint Conference on Natural Language Processing of the AFNLP},
  year={2015}
}

@article{talmor2018web,
  title={The web as a knowledge-base for answering complex questions},
  author={Talmor, Alon and Berant, Jonathan},
  journal={arXiv preprint arXiv:1803.06643},
  year={2018}
}

@inproceedings{zhang2018variational,
  title={Variational reasoning for question answering with knowledge graph},
  author={Zhang, Yuyu and Dai, Hanjun and Kozareva, Zornitsa and Smola, Alexander and Song, Le},
  booktitle={Proceedings of the AAAI conference on artificial intelligence},
  volume={32},
  number={1},
  year={2018}
}

@article{yue2025does,
  title={Does reinforcement learning really incentivize reasoning capacity in llms beyond the base model?},
  author={Yue, Yang and Chen, Zhiqi and Lu, Rui and Zhao, Andrew and Wang, Zhaokai and Song, Shiji and Huang, Gao},
  journal={arXiv preprint arXiv:2504.13837},
  year={2025}
}

@article{chu2025sft,
  title={Sft memorizes, rl generalizes: A comparative study of foundation model post-training},
  author={Chu, Tianzhe and Zhai, Yuexiang and Yang, Jihan and Tong, Shengbang and Xie, Saining and Schuurmans, Dale and Le, Quoc V and Levine, Sergey and Ma, Yi},
  journal={arXiv preprint arXiv:2501.17161},
  year={2025}
}

@inproceedings{li2024entropic,
  title={Entropic distribution matching for supervised fine-tuning of LLMs: Less overfitting and better diversity},
  author={Li, Ziniu and Chen, Congliang and Xu, Tian and Qin, Zeyu and Xiao, Jiancong and Sun, Ruoyu and Luo, Zhi-Quan},
  booktitle={NeurIPS 2024 Workshop on Fine-Tuning in Modern Machine Learning: Principles and Scalability},
  year={2024}
}

@inproceedings{liu2023lazy,
  title={Lazy agents: A new perspective on solving sparse reward problem in multi-agent reinforcement learning},
  author={Liu, Boyin and Pu, Zhiqiang and Pan, Yi and Yi, Jianqiang and Liang, Yanyan and Zhang, Du},
  booktitle={International Conference on Machine Learning},
  pages={21937--21950},
  year={2023},
  organization={PMLR}
}

@article{rengarajan2022reinforcement,
  title={Reinforcement learning with sparse rewards using guidance from offline demonstration},
  author={Rengarajan, Desik and Vaidya, Gargi and Sarvesh, Akshay and Kalathil, Dileep and Shakkottai, Srinivas},
  journal={arXiv preprint arXiv:2202.04628},
  year={2022}
}

@article{wang2025spa,
  title={SPA-RL: Reinforcing LLM Agents via Stepwise Progress Attribution},
  author={Wang, Hanlin and Leong, Chak Tou and Wang, Jiashuo and Wang, Jian and Li, Wenjie},
  journal={arXiv preprint arXiv:2505.20732},
  year={2025}
}

@article{deng2024novice,
  title={From novice to expert: Llm agent policy optimization via step-wise reinforcement learning},
  author={Deng, Zhirui and Dou, Zhicheng and Zhu, Yutao and Wen, Ji-Rong and Xiong, Ruibin and Wang, Mang and Chen, Weipeng},
  journal={arXiv preprint arXiv:2411.03817},
  year={2024}
}

@article{choudhury2025process,
  title={Process reward models for llm agents: Practical framework and directions},
  author={Choudhury, Sanjiban},
  journal={arXiv preprint arXiv:2502.10325},
  year={2025}
}

@article{yang2025qwen3,
  title={Qwen3 technical report},
  author={Yang, An and Li, Anfeng and Yang, Baosong and Zhang, Beichen and Hui, Binyuan and Zheng, Bo and Yu, Bowen and Gao, Chang and Huang, Chengen and Lv, Chenxu and others},
  journal={arXiv preprint arXiv:2505.09388},
  year={2025}
}

@article{dubey2024llama,
  title={The llama 3 herd of models},
  author={Dubey, Abhimanyu and Jauhri, Abhinav and Pandey, Abhinav and Kadian, Abhishek and Al-Dahle, Ahmad and Letman, Aiesha and Mathur, Akhil and Schelten, Alan and Yang, Amy and Fan, Angela and others},
  journal={arXiv e-prints},
  pages={arXiv--2407},
  year={2024}
}

@inproceedings{lightman2023let,
  title={Let's verify step by step},
  author={Lightman, Hunter and Kosaraju, Vineet and Burda, Yuri and Edwards, Harrison and Baker, Bowen and Lee, Teddy and Leike, Jan and Schulman, John and Sutskever, Ilya and Cobbe, Karl},
  booktitle={The Twelfth International Conference on Learning Representations},
  year={2023}
}

@article{paolo2024discovering,
  title={Discovering and exploiting sparse rewards in a learned behavior space},
  author={Paolo, Giuseppe and Coninx, Miranda and Laflaqui{\`e}re, Alban and Doncieux, Stephane},
  journal={Evolutionary Computation},
  volume={32},
  number={3},
  pages={275--305},
  year={2024},
  publisher={MIT Press 255 Main Street, 9th Floor, Cambridge, Massachusetts 02142, USA~…}
}

@inproceedings{su2025racqc,
  title={RACQC: Advanced Retrieval-Augmented Generation for Chinese Query Correction},
  author={Su, Jinbo and Gao, Lingzhe and Li, Wei and Liu, Shihao and Lei, Haojie and Wang, Xinyi and Guo, Yuanzhao and Wang, Ke and Shi, Daiting and Yin, Dawei},
  booktitle={Findings of the Association for Computational Linguistics: EMNLP 2025},
  pages={675--689},
  year={2025}
}

@misc{li2026sesearchselfevolvingsearchagent,
      title={SE-Search: Self-Evolving Search Agent via Memory and Dense Reward}, 
      author={Jian Li and Yizhang Jin and Dongqi Liu and Hang Ding and Jiafu Wu and Dongsheng Chen and Yunhang Shen and Yulei Qin and Ying Tai and Chengjie Wang and Xiaotong Yuan and Yabiao Wang},
      year={2026},
      eprint={2603.03293},
      archivePrefix={arXiv},
      primaryClass={cs.CL},
      url={https://arxiv.org/abs/2603.03293}, 
}

@misc{zhou2026arkanswercentricretrievertuning,
      title={ARK: Answer-Centric Retriever Tuning via KG-augmented Curriculum Learning}, 
      author={Jiawei Zhou and Hang Ding and Haiyun Jiang},
      year={2026},
      eprint={2511.16326},
      archivePrefix={arXiv},
      primaryClass={cs.IR},
      url={https://arxiv.org/abs/2511.16326}, 
}

@article{xu2026unlocking,
  title={Unlocking Implicit Experience: Synthesizing Tool-Use Trajectories from Text},
  author={Xu, Zhihao and Li, Rumei and Li, Jiahuan and Weng, Rongxiang and Wang, Jingang and Cai, Xunliang and Wang, Xiting},
  journal={arXiv preprint arXiv:2601.10355},
  year={2026}
}

@misc{xu2025videosegr1reasoningvideoobjectsegmentation,
      title={VideoSeg-R1:Reasoning Video Object Segmentation via Reinforcement Learning}, 
      author={Zishan Xu and Yifu Guo and Yuquan Lu and Fengyu Yang and Junxin Li},
      year={2025},
      eprint={2511.16077},
      archivePrefix={arXiv},
      primaryClass={cs.CV},
      url={https://arxiv.org/abs/2511.16077}, 
}

\clearpage

\appendix
\section{Training Details}

\label{sec:details}
\textbf{Implementation Details.} 
In our experiments, we primarily employed the Llama3.1-8B~\citep{dubey2024llama} and Qwen3-8B~\citep{yang2025qwen3} models. Our data synthesis pipeline produced 9,035 SFT and 3,504 RL training instances; with this data, the Qwen3-8B model was trained for 3 SFT and 15 RL epochs on 8 A100 GPUs, requiring 32 hours in total. 

For data generation, we apply our proposed data synthesis pipeline, producing a total of 9,035 training instances for SFT and 3,504 instances for RL.  
In the SFT stage, we fine-tune the Qwen3-8B with a learning rate of $1\times 10^{-4}$ for $3$ epochs.  
In the RL stage, we adopt the GRPO algorithm with a batch size of $512$, $15$ training epochs, a learning rate of $1\times 10^{-5}$, a value clipping range of $0.5$, and a KL divergence coefficient of $0.001$.  
The entire RL training phase takes approximately $32$ hours on $8$ NVIDIA A100 80GB GPUs.

\begin{table}[h]
\centering
\renewcommand{\arraystretch}{1.3} 
\setlength{\tabcolsep}{12pt}      
\begin{tabular}{ll}
\toprule
\textbf{Hyperparameter} & \textbf{Value} \\
\midrule
Learning rate       & $1\times 10^{-5}$ \\
Batch size          & 512 \\
Epochs              & 15 \\
Clip ratio          & 0.2 \\
Gradient clipping   & 1.0 \\
KL coefficient      & 0.001 \\
PPO mini-batch size & 16 \\
\bottomrule
\end{tabular}
\caption{Key hyperparameters for RL training (GRPO).}
\label{tab:rl_hparams}
\end{table}

\section{Prompt of Interactive Graph Retrieval}
\label{prompt}

\begin{tcolorbox}[
    title=\textbf{Prompts for Interactive Graph Retriever},
    colback=white,
    colframe=black!75,
    coltitle=white,
    fonttitle=\bfseries,
    boxrule=0.8pt,
    arc=2pt,
    fontupper=\small, 
    breakable,        
    enhanced,
    left=2pt, right=2pt, top=2pt, bottom=2pt 
]

You are an intelligent agent skilled in exploring Knowledge Graphs, with strong reasoning abilities. Your task is to perform question answering over a Knowledge Graph by gradually exploring it. You should start from the entities mentioned in the question and explore the graph step by step until you gather enough information to answer the question.

\vspace{0.5em}
\textbf{Your task follows these steps:}

\begin{enumerate}[leftmargin=1.2em, nosep] 
    \item \textbf{Understand the Question}
    
    \item \textbf{Analyze the Action History and Current Graph State}
    
    \item \textbf{Choose the Next Action} from the following options:
    \begin{itemize}[leftmargin=1em, topsep=2pt, itemsep=3pt]
        \item \textbf{"Explore Entity"}: Explore all triples directly connected to a given entity in the Knowledge Graph.
        
        \item \textbf{"Choose Relation"}: Select the triple(s) from the explored information that are most relevant to the question.\\
        \textit{Attention: Only the triples included in the "Objects" field of the "Choose Relation" step will be retained in the future "Current Graph State". So You must filter and retain the information useful for answering the question or for further exploration.}
        
        \item \textbf{"Finish"}: Choose this action when you believe you have gathered sufficient information to answer the question. Your final answers should be included in the "Objects" field.
    \end{itemize}

    \item \textbf{Select the Objects}: Depending on the action, provide the relevant entity or triple(s).\\
    \textit{Attention: All objects must come from the "Entities in Question" or the current "Current Graph State". Do not create new entities or relations.}
    
    \vspace{3pt}
    \textit{Examples:}
    \begin{itemize}[leftmargin=1em, nosep]
        \item If "Explore Entity": \texttt{"Objects": ["EntityA", "EntityB"]}
        \item If "Choose Relation": \texttt{"Objects": ["(Sub1, Rel1, Obj1)", ...]}
        \item If "Finish": \texttt{"Objects": ["Answer1", "Answer2"]}
    \end{itemize}

    \item Output your response in JSON format, and include a \textbf{detailed thought process} explaining your reasoning at this step.
\end{enumerate}

\noindent\rule{\linewidth}{0.4pt} 

\textbf{Question:}\\
\textbf{Entities in Question:}\\
\textbf{Current Graph State:}\\
\textbf{Action History:}

\noindent\rule{\linewidth}{0.4pt}

\vspace{0.5em}
Please respond using the following format:

\textbf{Thought Process:}\\
<Provide a step-by-step analysis>

\textbf{Action Decision:}
\begin{tcolorbox}[colback=gray!10, boxrule=0pt, frame hidden, sharp corners, left=2pt, right=2pt, top=2pt, bottom=2pt]
\ttfamily
```json\\
\{\\
\ \ "Action": "<The type of action: 'Explore Entity' | 'Choose Relation' | 'Finish'>",\\
\ \ "Objects": [<The entities or triples>]\\
\}\\
\end{tcolorbox}
\end{tcolorbox}

\section{Case Study}
\label{CaseStudy}
\begin{tcolorbox}[
    title=\textbf{Case 1}, 
    colback=white,
    colframe=black!75,
    coltitle=white,
    fonttitle=\bfseries,
    boxrule=0.8pt,
    arc=2pt,
    fontupper=\small, 
    breakable,        
    enhanced,
    left=2pt, right=2pt, top=2pt, bottom=2pt 
]
\textbf{Question:} When were the movies directed by [Sarah Polley] released? \\
\textbf{True Answer:} 2011, 2006, 2012 \\
\textbf{Action History:}
Step 1, Explore Entity, Objects: ['Sarah Polley'] \\
Step 2, Choose Relation, Objects: ['(Away from Her, directed\_by, Sarah Polley)', '(Stories We Tell, directed\_by, Sarah Polley)', '(Take This Waltz, directed\_by, Sarah Polley)'] \\
Step 3, Explore Entity, Objects: ['Stories We Tell', 'Away from Her', 'Take This Waltz'] \\
Step 4, Choose Relation, Objects: ['(Away from Her, release\_year, 2006)', '(Stories We Tell, release\_year, 2012)'] \\
Step 5, Explore Entity, Objects: ['Take This Waltz'] \\
Step 6, Choose Relation, Objects: ['(Take This Waltz, release\_year, 2011)'] \\
Step 7, Finish, Answer: ['2006', '2011', '2012'] \\
\textbf{Retrieved Triples:} (Away from Her, release\_year, 2006), (Away from Her, directed\_by, Sarah Polley), (Take This Waltz, release\_year, 2011), (Stories We Tell, directed\_by, Sarah Polley), (Stories We Tell, release\_year, 2012), (Take This Waltz, directed\_by, Sarah Polley) \\
\end{tcolorbox}

To demonstrate the efficacy of our interactive pruning mechanism against high-degree entities, we analyze the query: "When were the movies directed by [Sarah Polley] released?" The entity "Sarah Polley" presents a significant challenge as a "hub node" with dual roles: she is a prolific actress with numerous connections but has directed comparatively fewer films.

Graph-$S^3$ Success via Structure-Aware Pruning: As shown in the trajectory, Graph-$S^3$ effectively overcomes the noise of Sarah Polley's acting career. In Step 2, the agent leverages the "Choose Relation" action to strictly select triples containing the "directed\_by" relation, while deliberately pruning the dense subgraph of movies where she appears merely as an actress. This aligns with our framework's design to "avoid introducing redundant context into the LLM" and demonstrates the value of stepwise supervision in learning to distinguish task-relevant paths from high-degree noise. Consequently, the agent precisely navigates to the correct release years.

In contrast, baseline methods exhibited failures characteristic of shallow embedding similarity. The standard RAG suffered from severe lexical confusion. It retrieved unrelated entities such as "Pollyanna" and "Polly of the Circus", erroneously matching the token "Polly" with "Polley." This confirms that similarity-based matching often yields noisy results by ignoring precise relational constraints. G-Retriever failed to filter the entity's large neighborhood, retrieving generic attributes instead of the specific directorial path. This illustrates the limitation of one-shot retrievers that retrieve large neighborhoods discarding the graph's relational structure, whereas Graph-$S^3$'s interactive policy successfully identified the sparse signal within a noisy subgraph.

\begin{tcolorbox}[
    title=\textbf{Case 2}, 
    colback=white,
    colframe=black!75,
    coltitle=white,
    fonttitle=\bfseries,
    boxrule=0.8pt,
    arc=2pt,
    fontupper=\small, 
    breakable,        
    enhanced,
    left=2pt, right=2pt, top=2pt, bottom=2pt 
]

\textbf{Question:} The screenwriter [Jane Anderson] co-wrote movies with whom? \\
\textbf{True Answer:} Anne Heche, Terry Ryan \\
\textbf{Action History:} \\
Step 1, Explore Entity, Objects: ['Jane Anderson'] \\
Step 2, Choose Relation, Objects: ['(If These Walls Could Talk 2, written\_by, Jane Anderson)', '(Normal, written\_by, Jane Anderson)', '(The Prize Winner of Defiance, Ohio, written\_by, Jane Anderson)'] \\
Step 3, Explore Entity, Objects: ['If These Walls Could Talk 2', 'Normal', 'The Prize Winner of Defiance, Ohio'] \\
Step 4, Choose Relation, Objects: ['(If These Walls Could Talk 2, written\_by, Jane Anderson)', '(Normal, written\_by, Jane Anderson)', '(The Prize Winner of Defiance, Ohio, written\_by, Jane Anderson)']
Step 5, Finish, Answer: ['If These Walls Could Talk 2', 'Normal', 'The Prize Winner of Defiance, Ohio'] \\
\textbf{Retrieved Triples:} (The Prize Winner of Defiance, Ohio, written\_by, Jane Anderson), (Normal, written\_by, Jane Anderson), (If These Walls Could Talk 2, written\_by, Jane Anderson) \\

\end{tcolorbox}

To identify limitations in current reasoning policies, we analyze a failure case involving the collaborative query: "The screenwriter [Jane Anderson] co-wrote movies with whom?" The trajectory of the agent reveals a critical flaw in the processing of "sibling" relationships. In Steps 1 through 3, the agent correctly identified the intermediate nodes, successfully retrieving movies written by Jane Anderson.

The reasoning collapsed at Step 4. When expanding from the movie nodes to identify other writers, the agent exhibited a strong "self-verification bias." Instead of selecting edges pointing to new entities (the co-writers, such as Anne Heche), the agent selected the "written\_by" relations pointing back to the query entity, Jane Anderson. Consequently, instead of discovering the target "sibling" nodes, the agent entered a redundant verification loop. It merely reconfirmed that Jane Anderson wrote the movies and prematurely terminated the search, outputting the movie titles instead of the co-writers' names. The model assigns disproportionate attention weight to entities explicitly named in the query, causing it to overlook unmentioned but structurally relevant entities.

\section{An Example of The Reasoning Process of Graph-$S^3$}
\label{ReasoningProcess}

\begin{lstlisting}[
    language=python,
    backgroundcolor=\color{gray!3},
    basicstyle=\ttfamily\small\linespread{0.85}\selectfont\color{black},
    breaklines=true,
    frame=single,
    rulecolor=\color{gray!10},
    keywordstyle=\color{black}, 
    stringstyle=\color{black}, 
    commentstyle=\color{black}, 
    showstringspaces=false,
    framerule=0.8pt,
    framesep=5pt
]

{
    "step": 0,
    "question": "the films that share directors with the film [The Life of Oharu] were in which genres",
    "question_entities": [
        "The Life of Oharu"
    ],
    "now_state": [],
    "action_history": [],
    "extract_res": {
        "Action": "Explore Entity",
        "Objects": [
            "The Life of Oharu"
        ]
    }
},
{
    "step": 1,
    "question": "the films that share directors with the film [The Life of Oharu] were in which genres",
    "question_entities": [
        "The Life of Oharu"
    ],
    "now_state": [
        "(The Life of Oharu, starred_actors, Kinuyo Tanaka)",
        "(The Life of Oharu, directed_by, Kenji Mizoguchi)",
        "(The Life of Oharu, release_year, 1952)",
        "(The Life of Oharu, written_by, Kenji Mizoguchi)",
        "(The Life of Oharu, has_tags, kenji mizoguchi)",
        "(The Life of Oharu, in_language, Japanese)"
    ],
    "action_history": [
        "step 1, Explore Entity, Objects: ['The Life of Oharu']"
    ],
    "extract_res": {
        "Action": "Choose Relation",
        "Objects": [
            "(The Life of Oharu, directed_by, Kenji Mizoguchi)"
        ]
    }
},
{
    "step": 2,
    "question": "the films that share directors with the film [The Life of Oharu] were in which genres",
    "question_entities": [
        "The Life of Oharu"
    ],
    "now_state": [
        "(The Life of Oharu, directed_by, Kenji Mizoguchi)"
    ],
    "action_history": [
        "step 1, Explore Entity, Objects: ['The Life of Oharu']",
        "step 2, Choose Relation, Objects: ['(The Life of Oharu, directed_by, Kenji Mizoguchi)']"
    ],
    "extract_res": {
        "Action": "Explore Entity",
        "Objects": [
            "Kenji Mizoguchi"
        ]
    }
},
{
    "step": 3,
    "question": "the films that share directors with the film [The Life of Oharu] were in which genres",
    "question_entities": [
        "The Life of Oharu"
    ],
    "now_state": [
        "(Ugetsu, directed_by, Kenji Mizoguchi)",
        "(Sansho the Bailiff, directed_by, Kenji Mizoguchi)",
        "(The Crucified Lovers, directed_by, Kenji Mizoguchi)",
        "(The 47 Ronin, directed_by, Kenji Mizoguchi)",
        "(The Life of Oharu, directed_by, Kenji Mizoguchi)",
        "(Osaka Elegy, directed_by, Kenji Mizoguchi)",
        "(The Life of Oharu, written_by, Kenji Mizoguchi)",
        "(Sisters of the Gion, directed_by, Kenji Mizoguchi)",
        "(The Life of Oharu, has_tags, kenji mizoguchi)",
        "(Utamaro and His Five Women, directed_by, Kenji Mizoguchi)"
    ],
    "action_history": [
        "step 1, Explore Entity, Objects: ['The Life of Oharu']",
        "step 2, Choose Relation, Objects: ['(The Life of Oharu, directed_by, Kenji Mizoguchi)']",
        "step 3, Explore Entity, Objects: ['Kenji Mizoguchi']"
    ],
    "extract_res": {
        "Action": "Choose Relation",
        "Objects": [
            "(Ugetsu, directed_by, Kenji Mizoguchi)",
            "(Sansho the Bailiff, directed_by, Kenji Mizoguchi)",
            "(The Crucified Lovers, directed_by, Kenji Mizoguchi)",
            "(The 47 Ronin, directed_by, Kenji Mizoguchi)",
            "(Osaka Elegy, directed_by, Kenji Mizoguchi)",
            "(Sisters of the Gion, directed_by, Kenji Mizoguchi)",
            "(Utamaro and His Five Women, directed_by, Kenji Mizoguchi)"
        ]
    }
},
{
    "step": 4,
    "question": "the films that share directors with the film [The Life of Oharu] were in which genres",
    "question_entities": [
        "The Life of Oharu"
    ],
    "now_state": [
        "(Ugetsu, directed_by, Kenji Mizoguchi)",
        "(Sansho the Bailiff, directed_by, Kenji Mizoguchi)",
        "(The Crucified Lovers, directed_by, Kenji Mizoguchi)",
        "(The 47 Ronin, directed_by, Kenji Mizoguchi)",
        "(The Life of Oharu, directed_by, Kenji Mizoguchi)",
        "(Osaka Elegy, directed_by, Kenji Mizoguchi)",
        "(Sisters of the Gion, directed_by, Kenji Mizoguchi)",
        "(Utamaro and His Five Women, directed_by, Kenji Mizoguchi)"
    ],
    "action_history": [
        "step 1, Explore Entity, Objects: ['The Life of Oharu']",
        "step 2, Choose Relation, Objects: ['(The Life of Oharu, directed_by, Kenji Mizoguchi)']",
        "step 3, Explore Entity, Objects: ['Kenji Mizoguchi']",
        "step 4, Choose Relation, Objects: ['(Ugetsu, directed_by, Kenji Mizoguchi)', '(Sansho the Bailiff, directed_by, Kenji Mizoguchi)', '(The Crucified Lovers, directed_by, Kenji Mizoguchi)', '(The 47 Ronin, directed_by, Kenji Mizoguchi)', '(Osaka Elegy, directed_by, Kenji Mizoguchi)', '(Sisters of the Gion, directed_by, Kenji Mizoguchi)', '(Utamaro and His Five Women, directed_by, Kenji Mizoguchi)']"
    ],
    "extract_res": {
        "Action": "Explore Entity",
        "Objects": [
            "Ugetsu",
            "Sansho the Bailiff",
            "The Crucified Lovers",
            "The 47 Ronin",
            "Osaka Elegy",
            "Sisters of the Gion",
            "Utamaro and His Five Women"
        ]
    }
},
{
    "step": 5,
    "question": "the films that share directors with the film [The Life of Oharu] were in which genres",
    "question_entities": [
        "The Life of Oharu"
    ],
    "true_answer": [
        "Drama"
    ],
    "now_state": [
        "(Ugetsu, directed_by, Kenji Mizoguchi)",
        "(Sansho the Bailiff, directed_by, Kenji Mizoguchi)",
        "(The Crucified Lovers, directed_by, Kenji Mizoguchi)",
        "(The 47 Ronin, directed_by, Kenji Mizoguchi)",
        "(The Life of Oharu, directed_by, Kenji Mizoguchi)",
        "(Osaka Elegy, directed_by, Kenji Mizoguchi)",
        "(Sisters of the Gion, directed_by, Kenji Mizoguchi)",
        "(Sisters of the Gion, has_genre, Drama)",
        "(Ugetsu, has_genre, Drama)",
        "(Ugetsu, starred_actors, Masayuki Mori)",
        "(Utamaro and His Five Women, directed_by, Kenji Mizoguchi)"
    ],
    "action_history": [
        "step 1, Explore Entity, Objects: ['The Life of Oharu']",
        "step 2, Choose Relation, Objects: ['(The Life of Oharu, directed_by, Kenji Mizoguchi)']",
        "step 3, Explore Entity, Objects: ['Kenji Mizoguchi']",
        "step 4, Choose Relation, Objects: ['(Ugetsu, directed_by, Kenji Mizoguchi)', '(Sansho the Bailiff, directed_by, Kenji Mizoguchi)', '(The Crucified Lovers, directed_by, Kenji Mizoguchi)', '(The 47 Ronin, directed_by, Kenji Mizoguchi)', '(Osaka Elegy, directed_by, Kenji Mizoguchi)', '(Sisters of the Gion, directed_by, Kenji Mizoguchi)', '(Utamaro and His Five Women, directed_by, Kenji Mizoguchi)']",
        "step 5, Explore Entity, Objects: ['Ugetsu', 'Sansho the Bailiff', 'The Crucified Lovers', 'The 47 Ronin', 'Osaka Elegy', 'Sisters of the Gion', 'Utamaro and His Five Women']"
    ],
    "extract_res": {
        "Action": "Choose Relation",
        "Objects": [
            "(Ugetsu, has_genre, Drama)",
            "(Sisters of the Gion, has_genre, Drama)"
        ]
    }
},
{
    "step": 13,
    "question": "the films that share directors with the film [The Life of Oharu] were in which genres",
    "question_entities": [
        "The Life of Oharu"
    ],
    "true_answer": [
        "Drama"
    ],
    "now_state": [
        "(Ugetsu, directed_by, Kenji Mizoguchi)",
        "(Sansho the Bailiff, directed_by, Kenji Mizoguchi)",
        "(The Crucified Lovers, directed_by, Kenji Mizoguchi)",
        "(The 47 Ronin, directed_by, Kenji Mizoguchi)",
        "(The Life of Oharu, directed_by, Kenji Mizoguchi)",
        "(Osaka Elegy, directed_by, Kenji Mizoguchi)",
        "(Sisters of the Gion, directed_by, Kenji Mizoguchi)",
        "(Sisters of the Gion, has_genre, Drama)",
        "(Ugetsu, has_genre, Drama)",
        "(Utamaro and His Five Women, directed_by, Kenji Mizoguchi)"
    ],
    "action_history": [
        "step 1, Explore Entity, Objects: ['The Life of Oharu']",
        "step 2, Choose Relation, Objects: ['(The Life of Oharu, directed_by, Kenji Mizoguchi)']",
        "step 3, Explore Entity, Objects: ['Kenji Mizoguchi']",
        "step 4, Choose Relation, Objects: ['(Ugetsu, directed_by, Kenji Mizoguchi)', '(Sansho the Bailiff, directed_by, Kenji Mizoguchi)', '(The Crucified Lovers, directed_by, Kenji Mizoguchi)', '(The 47 Ronin, directed_by, Kenji Mizoguchi)', '(Osaka Elegy, directed_by, Kenji Mizoguchi)', '(Sisters of the Gion, directed_by, Kenji Mizoguchi)', '(Utamaro and His Five Women, directed_by, Kenji Mizoguchi)']",
        "step 5, Explore Entity, Objects: ['Ugetsu', 'Sansho the Bailiff', 'The Crucified Lovers', 'The 47 Ronin', 'Osaka Elegy', 'Sisters of the Gion', 'Utamaro and His Five Women']",
        "step 6, Choose Relation, Objects: ['(Ugetsu, has_genre, Drama)', '(Sisters of the Gion, has_genre, Drama)']"
    ],
    "extract_res": {
        "Action": "Finish",
        "Objects": [
            "Drama"
        ]
    }
}

\end{lstlisting}

\end{document}